\documentclass[runningheads]{llncs}

 \usepackage{relsize}
\usepackage{array,ragged2e}
\usepackage{graphicx}
\usepackage{pack}
\usepackage[export]{adjustbox}
\usepackage{tabularx} 
\usepackage{array}
\newcolumntype{P}[1]{>{\centering\arraybackslash}p{#1}}
\NewDocumentCommand{\set}{o m}{%
  \IfNoValueTF{#1}
    {\{#2\}}
    {\{#1 \mid #2\}}%
}
\begin{document}
\title{Extraction of Hierarchical Functional Connectivity Components in human brain using Adversarial Learning}
%
\titlerunning{Adversarial Learning based hSCP}
%
\author{Dushyant Sahoo and Christos Davatzikos}
%
%
\institute{University of Pennsylvania\\
\email{sadu@seas.upenn.edu}\\
}
\maketitle              

\begin{abstract}
The estimation of sparse hierarchical components reflecting patterns of the brain's functional connectivity from rsfMRI data can contribute to our understanding of the brain's functional organization, and can lead to biomarkers of diseases. However, inter-scanner variations and other confounding factors pose a challenge to the robust and reproducible estimation of functionally-interpretable brain networks, and especially to reproducible biomarkers. Moreover, the brain is believed to be organized hierarchically, and hence single-scale decompositions miss this hierarchy. The paper aims to use current advancements in adversarial learning to estimate interpretable hierarchical patterns in the human brain using rsfMRI data, which are robust to ``adversarial effects'' such as inter-scanner variations. We write the estimation problem as a minimization problem and solve it using alternating updates. Extensive experiments on simulation and a real-world dataset show high reproducibility of the components compared to other well-known methods. 

\keywords{fMRI analysis  \and Matrix factorization \and Adversarial Learning.}
\end{abstract}

\section{Introduction}
There has been a lot of research on estimating interpretable components of functional connectivity of the brain. However, these components are often vulnerable to confounding variations,  herein referred to as ``adversary'' using ML language, such as inter-scanner and inter-protocol variations, and rsfMRI noise or irrelevant fluctuations, which can considerably reduce these components' reproducibility and hence their utility as biomarkers of diseases that disrupt functional connectivity. To address this limitation, in this paper, we introduce adversarial learning aiming to estimate components that are robust to such confounding variations. Moreover, we use an existing hierarchical decomposition approach \cite{9285290}, which attempts to capture the brain's multi-scale functional organization.  

Numerous methods have been developed for estimating interpretable components that capture the complexity of the human brain. Single-layer matrix factorization approaches, such as ICA \cite{smith2009correspondence} and NMF \cite{potluru2008group}, are not sufficient to capture these complexities, as these components are believed to have hierarchical organization \cite{doucet2011brain}. In order to construct hierarchical components that are interpretable, we extend previous work on hierarchical Sparse Connectivity Patterns (hSCPs) \cite{eavani2015identifying,9285290}. This approach's advantages are that it can extract sparse, overlapping, and hierarchical components, which are some of the desired properties that can enhance the understanding of the human brain's working. It also extracts subject-specific information of the components that is useful for capturing heterogeneity in the data.  

We introduce the Adversarial hSCPs (Adv. hSCPs) method to enhance the sparse component's robustness, which can improve its generalization performance. We formulate the problem as a bilevel matrix factorization problem and solve it using alternate minimization. Our method is motivated by recent advances in matrix factorization approaches that have used adversarial training to achieve the state of the art performances \cite{he2018adversarial,luo2020adversarial}.  In a nutshell, it is a minimax game, where the adversary perturbs model parameters to maximize or deteriorate our objective function, and in defense, we minimize the objective function. We discuss more about adversarial learning in the upcoming sections. 

\textbf{Outline:} We start by reviewing hSCP and adversarial learning. Then, in Section 3, we present our method Adversarial hSCP. In Section 4 that follows, we compare our method against existing methods on simulated and real data. Additional experiments and calculations are in the Appendix.

\textbf{Contributions:} We propose adversarial learning for extracting hierarchical Sparse Connectivity Components (hSCP). Empirical studies performed on simulations and real-world datasets demonstrate that our model can generate more reproducible components than other related methods. We also extract interpretable components using the HCP dataset.

\textbf{Notations and Conventions}: We follow the same notations as used in \cite{9285290}. $\mathbb{S}^{P \times P}_{++}$ is the set of symmetric positive definite matrices of size $P \times P$. $\textbf{A} \geq 0$ denotes all the elements of the matrix are greater than or equal to $0$. $\textbf{1}_P$ denotes $P \times P$ matrix with all elements equal to one. $\textbf{I}_P$ denotes $P \times P$ identity matrix.  $\mathbf{A} \circ \mathbf{B}$ is the element wise product between two matrices $\mathbf{A}$ and $\mathbf{B}$. 

\section{PRELIMINARIES}

\subsection{Introduction to hierarchical Sparse Connectivity Patterns}
\label{sec:hSCP}
Let $\mathbf{X}_t^i = [X_{t1}^i,X_{t1}^i,\ldots,X_{tP}^i]$ be the fMRI BOLD responses of $i^{th}$ subject measured on $P$ nodes at time $t$, for $t=1,\ldots,T$. The approach takes the correlation matrix for each subject $\mathbf{\Theta}^i \in \mathbb{S}^{P \times P}_{++}$ as an input where each correlation matrix stores pairwise correlation between the nodes. In hSCP, $\mathbf{\Theta}^i $ is decomposed into non-negative linear combination of sparse hierarchical components as: 
 \begin{align*}
\mathbf{\Theta}^i &\approx \mathbf{W}_1\mathbf{\Lambda}_1^i \mathbf{W}_1^\top \\ & \vdots \\ \mathbf{\Theta}^i &\approx \mathbf{W}_1\mathbf{W}_2 \ldots\mathbf{W}_K \mathbf{\Lambda}^i_K \mathbf{W}_K^\top\mathbf{W}_{K-1}^\top \ldots \mathbf{W}_1^\top ,
 \end{align*}
 where $K$ is depth of hierarchy, $P > k_1 > \ldots > k_K $, set $\mathcal{W} = {\set[\mathbf{W}_r]{r=1,\ldots, K}}$ stores information about sparse components shared across all subjects and set $\mathcal{D}= {\set[\mathbf{\Lambda}_r^i]{r=1,\ldots, K; i=1\ldots,S}}$ stores subject specific diagonal matrix with $\mathbf{\Lambda}^i_r \geq 0$.  In the above formulation, $k_r$ denotes number of components at $r^{th}$ level. To understand the decomposition, we look at a two-layer hierarchy, where $\mathbf{W}_1 \in \mathbb{R}^{P \times k_1}$ stores $k_1$ components at the first layer, which is the bottom-most level. If we look at second level, which is the layer above the first, the components are comprised of linear combination of components at first level i.e. $\mathbf{W}_1 \times \mathbf{W}_2 \in \mathbb{R}^{P \times k_2}$, where $k_2 < k_1$. This repeated multiplication by low dimensional matrix will result in the coarser components which are at the top of the hierarchy. $L_1$ constraint on $\mathbf{W}_s$ is imposed to make $\mathbf{W}_1$ sparse resulting in sparse components and constraining rest of $\mathbf{W}$s to force the components at next layer to comprise of a sparse non-negative linear combination of the previous one. hSCP solves the below optimization to estimate hierarchical patterns:
 \begin{equation}
\begin{aligned}
& \underset{\mathcal{W},\mathcal{D}}{\text{minimize}}
& & H(\mathcal{W},\mathcal{D},\mathcal{C}) =  \sum_{i=1}^{S} \sum_{r=1}^{K}\|\mathbf{\Theta}^i - (\prod_{j=1}^{r}\mathbf{W}_j)\mathbf{\Lambda}_r^i(\prod_{n=1}^{r}\mathbf{W}_n)^\top \|_F^2\\
& \text{subject to}
&&\|\mathbf{w}^r_l\|_1 < \lambda_r,\; \|\mathbf{w}^r_l\|_\infty \leq 1, \; \trace(\mathbf{\Lambda}_r^i) =1,  \\
&&&\mathbf{\Lambda}_r^i \geq 0, \; \forall i,r,l; \hspace{2em} \mathbf{W}_j \geq 0, \; j=2,\ldots,K,
\end{aligned}
   \label{problem:hscp}
 \end{equation}
 where $\mathcal{C} = {\set[\mathbf{\Theta}^i]{i=1,\ldots, S}}$, $l=1,\ldots,k_r, \; i=1,\ldots,S \; \text{and} \; r=1,\ldots,K$. $L_1$, $L_{\infty}$ and $\trace$ constraints help the problem to identify a decomposition which can provide reproducible components. More details can be found in the original paper \cite{9285290}.

\subsection{Adversarial Training}
A discriminative model is usually trained by minimizing the empirical expected loss over a function class $\mathcal{F}  = \{f_{\mathbf{v}} : \mathbf{v} \in \mathcal{V} \}$:
\begin{align*}
    \min_{\mathbf{v} \in \mathcal{V}} \frac{1}{n} \sum_{i=1}^n l(f_{\mathbf{v}}(\mathbf{x}_i),y_i),
\end{align*}
where $l(\cdot,\cdot)$ is a loss function, $f_{\mathbf{v}}$ is the output function, $\mathbf{x}_i$ is the feature vector and $y_i$ is the label. Several recent papers \cite{goodfellow2014explaining,moosavi2017universal} have revealed that adding adversarial noise to the sample defined for a sample $(\mathbf{x}, y)$ as:
\begin{align}
    \delta_{\mathbf{v}}^{adv}(\mathbf{x}) := \argmax_{\|\delta \| \leq \epsilon} l(f_{\mathbf{v}}(\mathbf{x}+\delta),y),
    \label{eq:adv_noise}
\end{align}
where $\epsilon > 0$ is the adversarial noise power can drastically decrease model's performance. Adversarial training \cite{madry2018towards} was introduced to provide robustness against the adversaries defined above. The training involves empirical risk minimization over the perturbed samples by solving 
\begin{align*}
    \min_{\mathbf{v} \in \mathcal{V}} \frac{1}{n} \sum_{i=1}^n l(f_{\mathbf{v}}(x_i+\delta_{\mathbf{v}}^{adv}(\mathbf{x})),y_i).
\end{align*}
The above formulation has a drawback that the accuracy drops drastically. To overcome this problem, Mix-minibatch adversarial training (MAT) \cite{wong2018provable} is performed by solving
\begin{align}
    \min_{\mathbf{v} \in \mathcal{V}} \frac{1}{n} \sum_{i=1}^n l(f_{\mathbf{v}}(x_i+\delta_{\mathbf{v}}^{adv}(\mathbf{x})),y_i) +  l(f_{\mathbf{v}}(x_i),y_i),
    \label{eq:disc.adv}
\end{align}
which balances between accuracy on the clean examples and robustness on the adversarial examples. Motivated by the above methodology, we build adversarial training for learning sparse hierarchical connectivity components. As the above training regime was supervised, we use a different formulation with the same idea for the unsupervised hSCP model.

\section{Adversarial hierarchical Sparse Connectivity Patterns}
In this section, we demonstrate how to incorporate adversarial learning at one level for the hSCP method, which then can be extended to multiple levels. The idea is to perturb input data $\mathbf{\Theta}^i$ and learn stable components $\mathbf{W}_1$ which are robust to adversaries, such as inter-scanner differences and unwanted rsfMRI noise. There are two parts of the complete learning procedure-

\textbf{Attack}- We first manually perturb the input  data to get perturbed data $\mathbf{\Gamma}^i = \mathbf{\Theta}^i + 0.1\sigma \mathbf{1}_P$ where $\sigma$ is the standard deviation of the data and use it to learn a perturbed weight matrix $\mathbf{\tilde{W}}_1$ by minimizing the below cost function:
    \begin{align}
       A(\mathbf{\hat{W}}_1) = \alpha\| \mathbf{\tilde{W}}_1 - \mathbf{W}_1\|_F^2 +  \sum_{i=1}^{S} \|\mathbf{\Gamma}^i  - \mathbf{\tilde{W}}_1 \mathbf{\Lambda}^i \mathbf{\tilde{W}}^\top_1 \|_F^2.
       \label{eq:attack}
    \end{align}
 In the above equation, the first part is used to estimate $\mathbf{\tilde{W}}_1$, which is close to $\mathbf{{W}}_1$ in Frobenius norm, thus mimicking the actual components but is learned from the noise-induced data. The second term is used for learning $\mathbf{\tilde{W}}_1$ using a perturbed data matrix. The main goal of the attacker is to learn $\mathbf{\tilde{W}}_1$ for a given $\mathbf{\Gamma}^i$ and fool the model by forcing the model to learn $\mathbf{\Lambda}^i$ from the perturbed data. Our framework does not depend on the perturbations' assumptions; the type of perturbation can be varied depending on different types of practical noises such as site-induced or scanner-induced noise. Finding optimal perturbation is left for future work.

 \textbf{Defense}- Aim of the learner is to estimate $\mathbf{\Lambda}^i$ and $\mathbf{W}_1$ by minimizing the below cost function:
    \begin{align}
     D(\mathbf{W},\mathbf{\Lambda}) =    \sum_{i=1}^{S} \|\mathbf{\Theta}^i  - \mathbf{\tilde{W}}_1\mathbf{\Lambda}^i \mathbf{\tilde{W}}^\top_1 \|_F^2 +\beta \sum_{i=1}^{S}\|\mathbf{\Theta}^i  - \mathbf{W}_1\mathbf{\Lambda}^i \mathbf{{W}}^\top_1 \|_F^2 ,
     \label{eq:defense}
    \end{align}
    for a fixed $\mathbf{\tilde{W}}_1$. Learner first estimates subject specific information $\mathbf{\Lambda}^i$ using perturbed weight matrix and then use it to learn $\mathbf{{W}}_1$. We can now define the optimization problem for the complete adversarial learning at single level using equation \ref{eq:attack} and \ref{eq:defense} as:
 \begin{equation}
\begin{aligned}
& \underset{\mathbf{W}_1,\mathbf{\Lambda}^i \; \forall i}{\text{minimize}}
& &  \sum_{i=1}^{S} \|\mathbf{\Theta}^i  - \mathbf{\tilde{W}}_1\mathbf{\Lambda}^i \mathbf{\tilde{W}}^\top_1 \|_F^2 + \beta \sum_{i=1}^{S}\|\mathbf{\Theta}^i  - \mathbf{W}_1\mathbf{\Lambda}^i \mathbf{{W}}^\top_1 \|_F^2\\
& \text{subject to}
&&\mathbf{\tilde{W}}_1 = \argminA_{\mathbf{\hat{W}}_1} \alpha\| \mathbf{\hat{W}}_1 - \mathbf{{W}}_1\|_F^2 + \sum_{i=1}^{S} \|\mathbf{\Gamma}^i  - \mathbf{\tilde{W}}_1 \mathbf{\Lambda}^i \mathbf{\tilde{W}}^\top_1 \|_F^2
\end{aligned}
   \label{problem:adv}
 \end{equation}
The above equation is analogous to discriminate adversarial learning problem defined in equation \ref{eq:disc.adv}. Multi level formulation of the above problem can be written as: 
  \begin{equation}
\begin{aligned}
& \underset{\mathcal{W},\mathcal{D}}{\text{minimize}}
& & J(\mathcal{\tilde{W}},\mathcal{W},\mathcal{D},\mathcal{C})= H(\mathcal{\tilde{W}},\mathcal{D},\mathcal{C}) + \beta H(\mathcal{W},\mathcal{D},\mathcal{C}) \\
& \text{subject to}
&& \mathbf{\tilde{W}}_r = \argminA_{\mathbf{\hat{W}}_r} \alpha\| \mathbf{\hat{W}}_r - \mathbf{{W}}_r\|_F^2 + H(\mathcal{\tilde{W}},\mathcal{D},\mathcal{P}) \quad r=1,\ldots,K
\end{aligned}
   \label{problem:h.adv}
 \end{equation}
where $\mathcal{\tilde{W}} = {\set[\mathbf{\tilde{W}}_r]{r=1,\ldots, K}}$ and $\mathcal{P} = {\set[\mathbf{\Gamma}^i]{i=1,\ldots, S}}$.

\begin{algorithm}[t]
 \caption{Adv. hSCP}\label{alg:adv.hscp}
\begin{algorithmic}[1]
\State \textbf{Input:} Data $\mathcal{C}$, perturbed data $\mathcal{P}$; $\mathcal{W}$ and $\mathcal{D}$ = hSCP($\mathcal{C}$) 
\Repeat 
\For{$r=1$ {\bfseries to} $K$} 
\State \textit{Update adversarial perturbations}
\State $\mathbf{\hat{W}}_r \leftarrow \descent(\mathbf{\hat{W}}_r)$
\State \textit{Update model parameters}
\State $\mathbf{W}_r \leftarrow \descent(\mathbf{W}_r)$
\If{$r==1$}
   \State $\mathbf{W}_r \leftarrow \proj_1(\mathbf{W}_r) $
   \Else{}
\State $\mathbf{W}_r \leftarrow \proj_2(\mathbf{W}_r) $
   \EndIf
\State $\mathbf{\Lambda}_r^i \leftarrow \descent(\mathbf{\Lambda}_r^i)$; $\mathbf{\Lambda}_r^i \leftarrow \proj_2(\mathbf{\Lambda}_r^i) \quad i=1,\ldots,S $
\EndFor
\Until{Stopping criterion is reached}
\State \textbf{Output:} $\mathbf{W}$ and $\mathbf{\Lambda}$
\end{algorithmic}
\end{algorithm} 

\subsection{Optimization}
The complete algorithm to solve the above optimization problem is described in Algorithm \ref{alg:adv.hscp} (Adv. hSCP). First, the adversarial perturbations are generated by performing gradient descent on $\mathbf{\hat{W}}_r$, and then the model parameters are updated using gradient descent. This process is repeated until the convergence criteria is reached. $\descent$ is the update rules defined by AMSgrad \cite{reddi2019convergence} for performing gradient descent. The update rule and gradients are defined in Appendix. $\proj_1(\mathbf{A})$ \cite{podosinnikova2013robust} function projects each column of $\mathbf{A}$ into intersection of $L_1$ and $L_\infty$ ball defined in equation \ref{problem:hscp} and $\proj_2(\mathbf{A})$ function makes all the negative elements of $\mathbf{A}$ equal to zero. The model parameters are initialized by first optimizing the hSCP model using Algorithm 1 (hSCP) defined in \cite{9285290}, rather than randomly initialized. This makes algorithm deterministic, and the algorithm can start from an optimal point on which adversarial learning can improve if there is overfitting. 

\section{Experiments}

\subsection{Comparison with existing methods}
\subsubsection{Simulated dataset} 
\begin{table}[t!]
\caption{Accuracy on simulated dataset}\label{table:simu}
\centering
        \begin{tabular}{ m{2cm}P{2.3cm}P{2.3cm}P{2.3cm}P{2.3cm}  }
 \toprule
 \multicolumn{1}{c}{Method} & \multicolumn{1}{c}{$k_1=6$} & \multicolumn{1}{c}{$k_1=8   $}  & \multicolumn{1}{c}{$k_1=10$} & \multicolumn{1}{c}{$k_1=12$} \\
 \midrule
hSCP & $0.801$    & $0.829$ &   $0.818$ & $0.814$\\
Adv. hSCP & $0.832 $     & $0.847$ &  $0.867$  & $0.864$ \\
ICA & $0.656 \pm 0.004 $     & $0.696 \pm 0.022$ &  $0.734 \pm 0.025$  & $0.748 \pm 0.011$ \\
NMF & $0.650 \pm 0.104 $     &$0.701 \pm 0.071$  &  $0.708 \pm 0.118$  &  $0.712 \pm 0.085$\\
Adv. NMF & $0.695 \pm 0.047 $     & $0.718 \pm 0.069$ &  $0.720 \pm 0.091$  & $0.723 \pm 0.114$ \\
 \bottomrule
\end{tabular}
\end{table}
\begin{table}[t!]  \captionof{table}{{Accuracy on simulated data with two level hierarchy}}\label{table:simu_hi}
\centering
  \begin{tabular}{m{1cm}p{1.7cm}P{1.5cm}P{1.5cm}P{1.5cm}P{1.5cm}}
    \toprule
    & & \multicolumn{1}{c}{$k_1=6$} & \multicolumn{1}{c}{$k_1=8$}  & \multicolumn{1}{c}{$k_1=10$} & \multicolumn{1}{c}{$k_1=12$} \\ 
    \midrule
        \multirow{ 2}{*}{$k_2=4$} & hSCP & $0.821$ &  $0.837$ &   $0.827$  &   $0.819$  \\   
    & Adv. hSCP & $0.859$  & $0.864$ & $0.846$    &   $0.823$   \\
    \midrule
        \multirow{ 2}{*}{$k_2=6$}& hSCP & $0.816$ &  $0.819$ &   $0.813$ &   $0.805$ \\
   &Adv hSCP & $0.848$  & $0.849$ & $0.826$   &   $0.814$     \\
    \bottomrule
  \end{tabular}
\end{table}
We first use a simulated dataset to evaluate the performance of our model against SCP \cite{eavani2015identifying}, NMF \cite{potluru2008group}, adv. NMF \cite{luo2020adversarial} and ICA \cite{smith2009correspondence}. The values of $\alpha$ and $\beta$ are set to be $10^{-3}$ and $0.5$ respectively throughout the paper. We generate sparse components $\mathbf{S}_1 \in \mathbb{R}^{P \times k_1}$ with $P = 50$ and $k_1 = 8$ and generate network structure from it. This network is then used as input to NetSim \cite{smith2011network} with TR equal to $3$ seconds to generate time-series data of $100$ subjects, each having $300$ time-points. NetSim also adds Gaussian noise to the time series of each node. We also add Poisson noise with a mean equal to $0.4$ to check how different methods perform in a high noise scenario.

We compare components/factors derived from all the models with $k_1 \in \{6,8,10,12 \}$. Accuracy is used as a performance measure defined as a normalized inner product between ground truth components and estimated factors derived from various algorithms. The optimal sparsity parameter $\lambda_1$ in the hSCP is selected from $ P\times 10^{[-2:1]}$ having the highest average split-sample reproducibility in $20$ runs. Split-sample reproducibility of components is computed by randomly dividing the data into two equal parts and then calculating normalized inner product between components extracted from each sample. A high reproducibility value implies that the same component can be extracted from multiple samples. Table \ref{table:simu} shows the comparison of the accuracy of different methods averaged over $20$ runs. As the hSCP method is deterministic, the output remains the same in every run. From the table, it can be seen that adversarial training can significantly improve the accuracy of hSCP. An important thing to note here is that adversarial training has also improved the accuracy of NMF, but it remains less than that of hSCP. The result for Poisson noise case is presented in Appendix.

We next generated a two-level hierarchy using the components defined above as the first layer. We used linear operator for projection to lower dimensional space to get coarse components with $P = 50$ and $k_2 = 4$. Visualization of the components is in Appendix. Time-series data were then generated under the same settings presented above. Table \ref{table:simu_hi} shows the components' average accuracy at two-level over $20$ runs for hSCP and Adv. hSCP with Adv. hSCP method giving better results. We did not use ICA and NMF as they can only generate components at only one level.

\subsubsection{Resting state fMRI data}
We used $100$ unrelated subjects released within the $900$ subjects data release from the publicly available Human Connectome Project (HCP) \cite{van2012human} dataset for comparing different methods. ICA+FIX pipeline \cite{glasser2013minimal} is used to process the complete data. Each subject has $4$ scans, with each scan comprising $1001$ time points and $360$ nodes.

We compare components/factors derived from all the models with $k_1 \in \{5,10,15,20 \}$. As the ground truth is not known, we use split-sample reproducibility as a performance measure. We first find the optimal value of $\lambda_1$ from $ P\times 10^{[-2:1]}$. Training, validation, and test data are generated by dividing data equally into three parts, and we then select $\lambda_1$ with the highest mean reproducibility over $20$ runs on training and test data. Training and test are used for final reproducibility comparison. Table \ref{table:hcp} shows that the hSCP method can extract components with high reproducibility. We have similar results presented in Table \ref{table:hcp_hi} for a two-level decomposition.

\begin{table}[t!]
\caption{Reproducibility on HCP dataset}\label{table:hcp}
\centering
        \begin{tabular}{ p{2cm}P{2.3cm}P{2.3cm}P{2.3cm}P{2.3cm}  }
 \toprule
Method & $10$ & $15$ & $20$ & $25$ \\
 \midrule
hSCP & $0.749 \pm  0.045$    & $0.750 \pm 0.046$ &   $0.712 \pm 0.026$ & $0.701 \pm 0.019$\\
Adv. hSCP & $0.787 \pm 0.052 $     & $0.765 \pm 0.059$ &  $0.716 \pm 0.020$  & $0.721 \pm 0.016$ \\
ICA & $0.695 \pm 0.067 $     & $0.638 \pm 0.046$ &  $0.581 \pm 0.039$  & $0.523 \pm 0.027$ \\
NMF & $0.689 \pm 0.038 $     &$0.657 \pm 0.067$  &  $0.635 \pm 0.053$  &  $0.629 \pm 0.020$\\
Adv. NMF & $0.709 \pm 0.073 $     & $0.659 \pm 0.043$ &  $0.653 \pm 0.026$  & $0.633 \pm 0.032$ \\
 \bottomrule
\end{tabular}
\end{table}
\begin{table}[t!]  \captionof{table}{{Reproducibility on HCP dataset with two level hierarchy}} \label{table:hcp_hi}
\centering
  \begin{tabular}{m{1cm}p{1.7cm}m{2.2cm}m{2.2cm}m{2.2cm}m{2.2cm}}
    \toprule
    & & \multicolumn{1}{c}{$k_1=10$} & \multicolumn{1}{c}{$k_1=15$}  & \multicolumn{1}{c}{$k_1=20$} & \multicolumn{1}{c}{$k_1=25$} \\ 
    \midrule
        \multirow{ 2}{*}{$k_2=4$} & hSCP & $0.872\pm 0.044$ &  $0.853\pm 0.064$ &   $0.831\pm 0.075$  &   $0.826\pm 0.091$  \\   
    & Adv. hSCP & $0.895\pm 0.030$  & $0.866\pm 0.029$ & $0.848\pm 0.056$    &   $0.830\pm 0.061$   \\
    \midrule
        \multirow{ 2}{*}{$k_2=6$}& hSCP & $0.856\pm 0.070$ &  $0.842\pm 0.062$ &   $0.828\pm 0.031$ &   $0.824\pm 0.035$ \\
   &Adv hSCP & $0.877\pm 0.076$  & $0.864\pm 0.067$ & $0.843\pm 0.045$   &   $0.834\pm 0.048$     \\
    \bottomrule
  \end{tabular}
\end{table}
\vspace{-1em}
\subsection{Results from rsfMRI data}
\begin{figure*}[tbp]
\begin{subfigure}{0.49\textwidth}
    \centering
\begin{minipage}[ct]{\textwidth}
\centering
\begin{forest}
  styleA/.style={top color=white, bottom color=white},
  styleB/.style={%
    top color=white,
    bottom color=red!20,
    delay={%
      content/.wrap value={##1\\{\includegraphics[scale=.5]{hand}}}
    }
  },
  for tree={
    rounded corners,
    draw,
    align=center,
    top color=white,
    bottom color=blue!20,
  },
  forked edges,
  [1{ \includegraphics[trim={1cm 3cm 1cm 2cm},clip,scale=0.15]{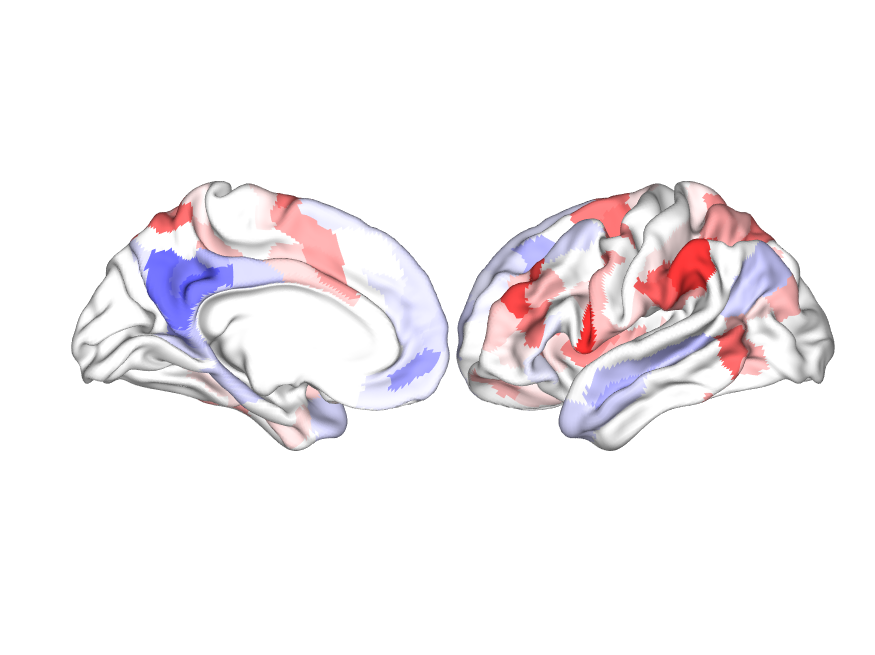}},styleA
    [{ \includegraphics[trim={1cm 3cm 1cm 2cm},clip,scale=0.15]{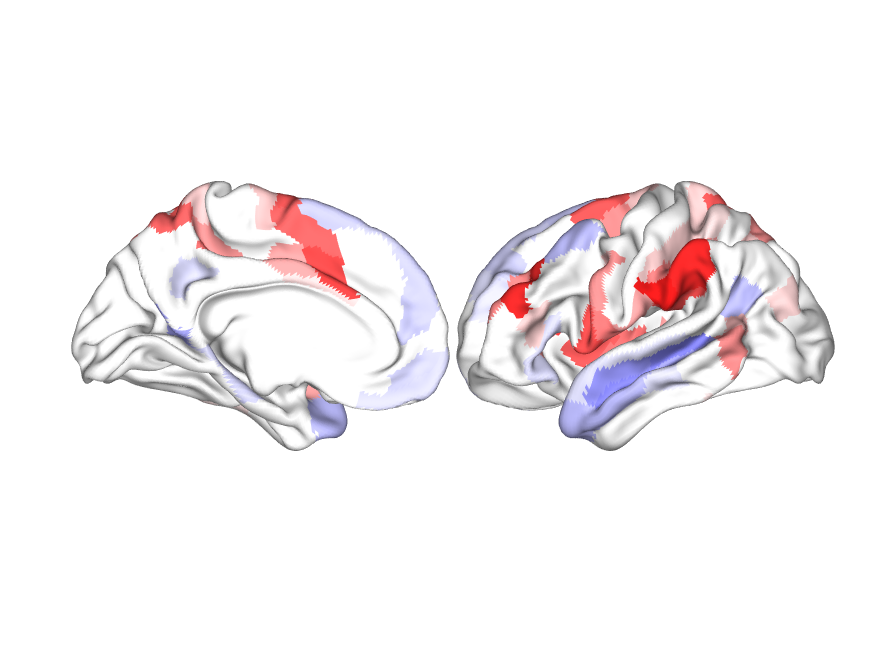}},styleA]
    [{ \includegraphics[trim={1cm 3cm 1cm 2cm},clip,scale=0.15]{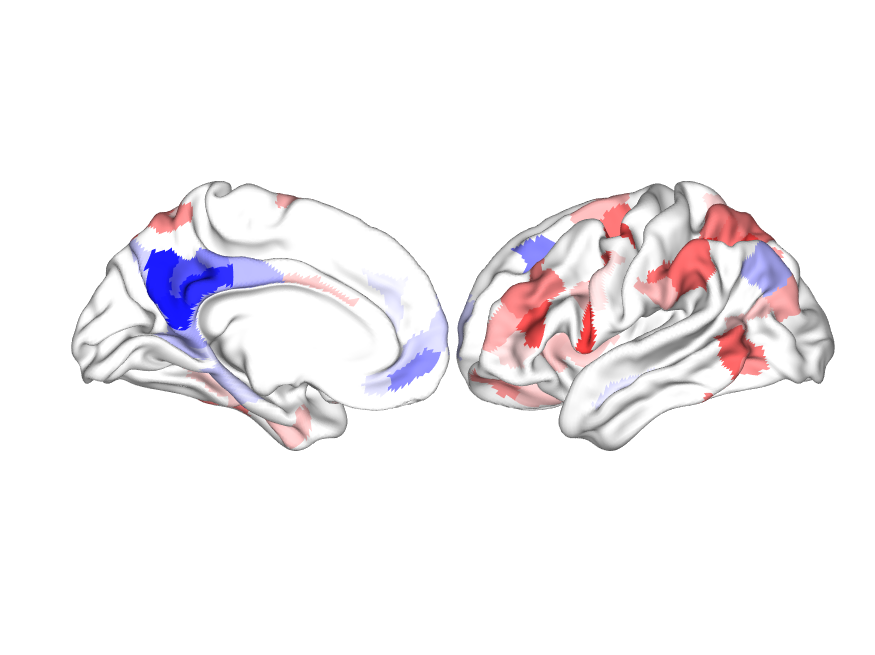}},styleA]
     ]
  ]
\end{forest}
\end{minipage}
\end{subfigure}
\vspace{1em}
\hfill
\begin{subfigure}{0.49\textwidth}
\begin{minipage}[ct]{\textwidth}
\centering
\begin{forest}
  styleA/.style={top color=white, bottom color=white},
  styleB/.style={%
    top color=white,
    bottom color=red!20,
    delay={%
      content/.wrap value={##1\\{\includegraphics[scale=.5]{hand}}}
    }
  },
  for tree={
    rounded corners,
    draw,
    align=center,
    top color=white,
    bottom color=blue!20,
  },
  forked edges,
[2{ \includegraphics[trim={1cm 3cm 1cm 2cm},clip,scale=0.15]{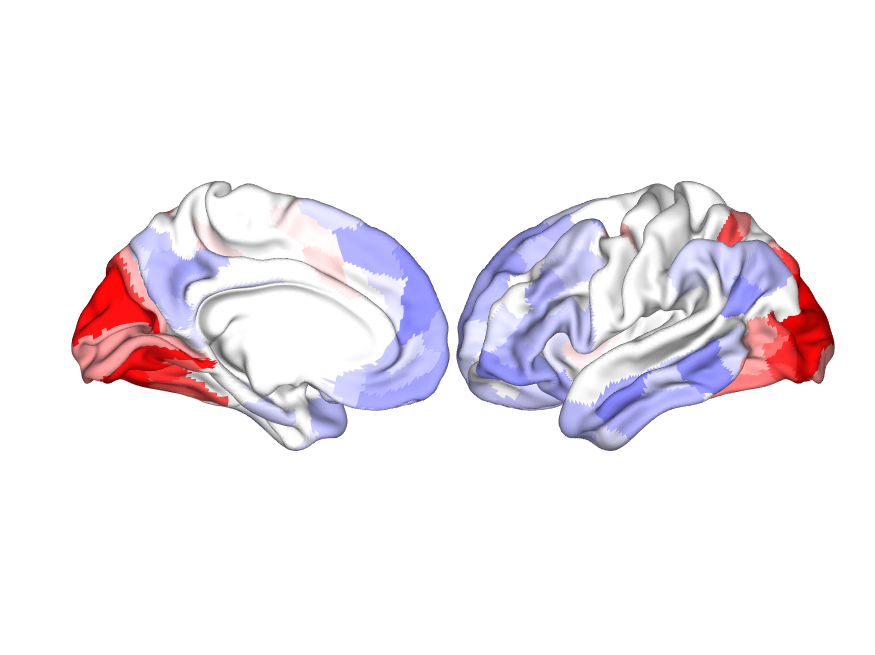}},styleA
    [{ \includegraphics[trim={1cm 3cm 1cm 2cm},clip,scale=0.15]{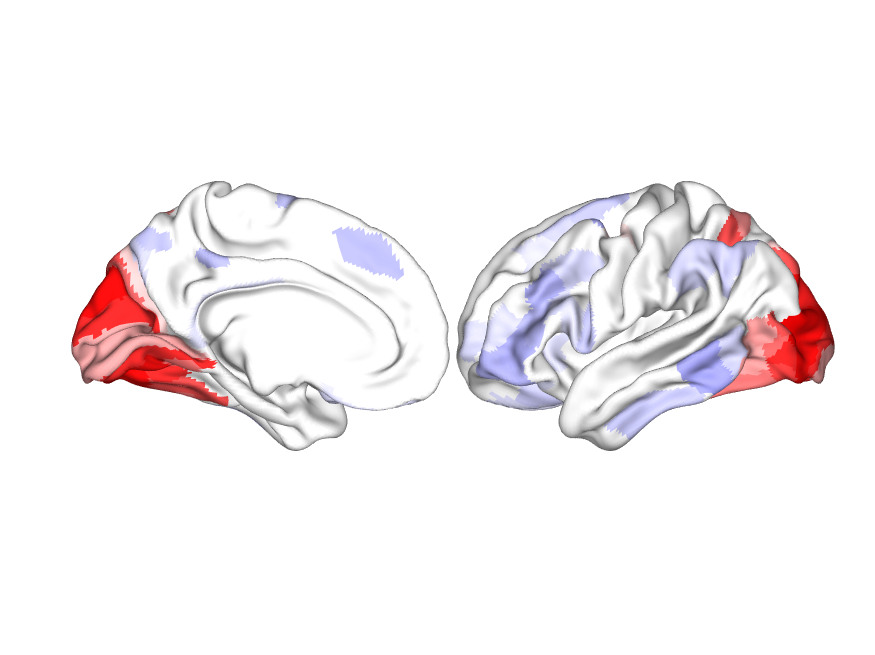}},styleA]
    [{ \includegraphics[trim={1cm 3cm 1cm 2cm},clip,scale=0.15]{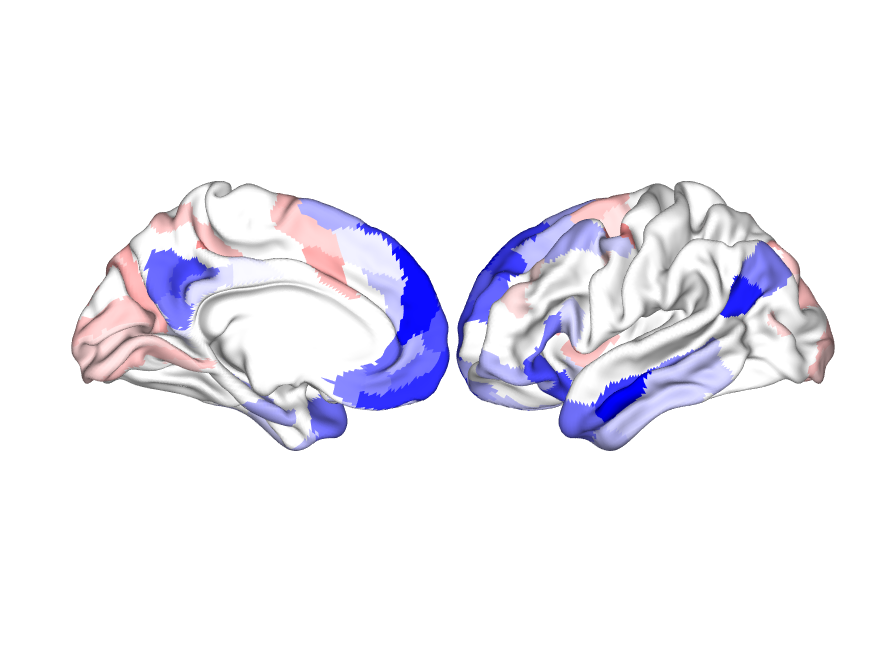}},styleA]
     ]
  ]
\end{forest}
\end{minipage}
\end{subfigure}
   \caption{ 
   Hierarchical components estimated using Adv. hSCP. Red and blue color are used for showing negative correlations between regions in a component.   }
    \label{fig:hir}
\end{figure*}

We extract $10$ components at level $1$, and $4$ components at level $2$ using Adv. hSCP learning from the HCP dataset. Figure \ref{fig:hir} shows two hierarchical components.  Component $1$ stores anti-correlation information between Default Mode Network and Dorsal Attention Network previously studied using seed-based correlation method \cite{fox2005human}. Component $2$ stores anti-correlation between Default Mode Network and extrastriate visual areas, which is another well-known finding \cite{uddin2009functional}. A more thorough discussion is needed for examining the differences and similarities between the components derived from hSCP and Adv. hSCP, which we have left for future work.  

\section{Conclusion}

In this study, we used adversarial learning to enhance the hSCP method by increasing the hierarchical components' reproducibility. We formulate the problem as a bilevel optimization problem and used adaptive gradient descent to solve it. Experimental results based on simulated data show that Adv hSCP can extract components accurately compared to other methods. Results using real-world rsfMRI data demonstrate the adversarial learning can improve the reproducibility of the components. We also discuss the interpretability of the components extracted from the HCP dataset.

There are several applications of this work. Improved reproducibility of the components can increase accuracy and confidence when applied to clinical applications such as age prediction, disease diagnosis, etc. Adversarial learning can be extended to other matrix factorization approaches used for the analysis of fMRI data, such as dynamic sparse connectivity patterns \cite{cai2017estimation}, sparse granger causality patterns \cite{sahoo2018gpu}, deep non-negative matrix factorization \cite{li2018identification}, etc. It would be interesting to assess the impact of the method in characterizing activity in terms of task-induced activations.

%
%

%
%
%
 \bibliographystyle{splncs04}
 \bibliography{main}

\newpage
\section{Appendix}
\vspace{-2em}
\begin{figure}[!htb]\centering
    \begin{subfigure}[t]{0.4\textwidth}
        \centering
        \includegraphics[width=\textwidth]{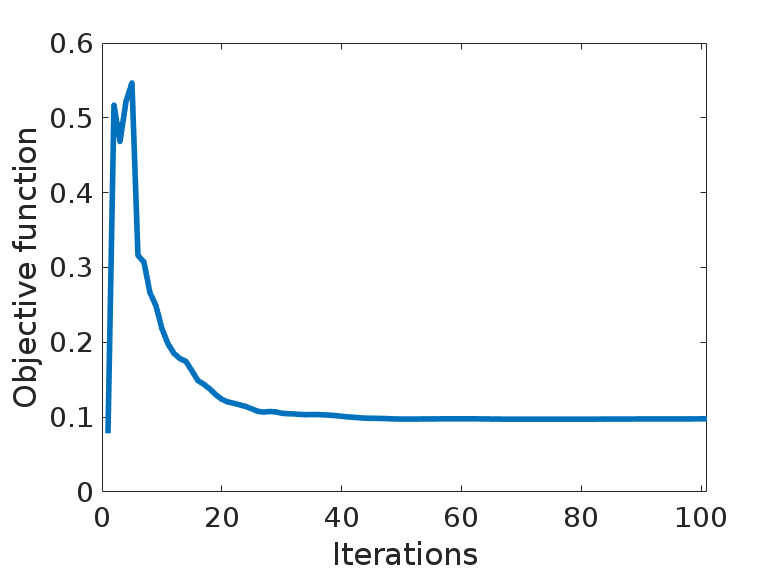}
        \caption{$k=10$}
    \end{subfigure}%
    \begin{subfigure}[t]{0.4\textwidth}
        \centering
        \includegraphics[width=1.1\textwidth]{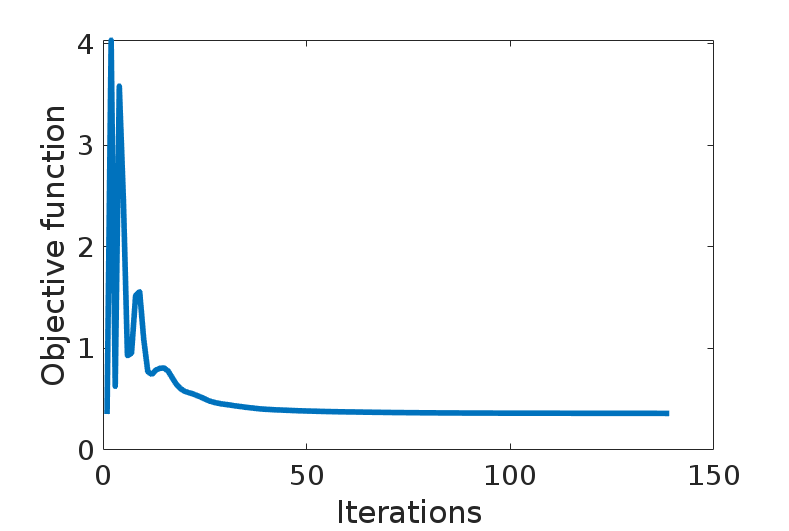}
        \caption{$k=20$ }
    \end{subfigure}
    \caption{Convergence of Adv. hSCP algorithm using HCP dataset for different values of $k$. In the figure, it can be seen that initially, the function value is low because the initial value is an optimal value of $\mathbf{W}$ and $\mathbf{\Lambda}$ returned using the hSCP algorithm. As the adversarial attack begins, the objective function value starts to fluctuate because of the minimax game, where the adversarial perturbation tries to deviate the result from the optimal value. In defense, we try to minimize the objective function. The algorithm converges when the optimal value becomes robust to perturbations.}
\end{figure}
\vspace{-3em}
\subsection{Gradients}
In this section, we define gradients used during adaptive gradient descent. Let
\begin{align*}
\mathbf{\tilde{W}}_0 &= \mathbf{W}_0 = \mathbf{I}_P, \qquad
\mathbf{Y}_r = \prod_{j=0}^{r}\mathbf{W}_j, \qquad \mathbf{\tilde{Y}}_r = \prod_{j=0}^{r}\mathbf{\tilde{W}}_j, \\ 
\mathbf{T}_{n,i}^r &= (\prod_{j=1}^{n-r}\mathbf{W}_j)\mathbf{\Lambda}^i_{n-r}(\prod_{j=1}^{n-r}\mathbf{W}_j)^\top, \qquad \mathbf{\tilde{T}}_{n,i}^r = (\prod_{j=1}^{n-r}\mathbf{\tilde{W}}_j)\mathbf{\Lambda}^i_{n-r}(\prod_{j=1}^{n-r}\mathbf{\tilde{W}}_j)^\top.
\end{align*}
We first define gradient for updating adversarial perturbations $\mathbf{\tilde{W}_r}$. The objective function is $F = \alpha\| \mathbf{\hat{W}}_r - \mathbf{{W}}_r\|_F^2 + H(\mathcal{\tilde{W}},\mathcal{D},\mathcal{P}) $ and gradient with respect to $\mathbf{\tilde{W}_r}$ will be
\begin{align*}
    \frac{F}{\partial \mathbf{\tilde{W}_r}} & = 2\alpha(\mathbf{\hat{W}}_r - \mathbf{{W}}_r) + \frac{\partial H(\mathcal{\tilde{W}},\mathcal{D},\mathcal{C})}{\partial \mathbf{\tilde{W}}_r}  = 2\alpha(\mathbf{\hat{W}}_r - \mathbf{{W}}_r)\\& \hspace{-02.5em}+  \sum_{i=1}^S\sum_{j=r}^{K} -4\mathbf{\tilde{Y}}_{r-1}^\top\mathbf{\Gamma}_i\mathbf{\tilde{Y}}_{r-1}\mathbf{\tilde{W}}_r\mathbf{\tilde{T}}_{j,i}^r  +   4\mathbf{\tilde{Y}}_{r-1}^\top\mathbf{\tilde{Y}}_{r-1}\mathbf{\tilde{W}}_r\mathbf{\tilde{T}}_{j,i}^r\mathbf{\tilde{W}}_r^\top\mathbf{\tilde{Y}}_{r-1}^\top\mathbf{\tilde{Y}}_{r-1}\mathbf{\tilde{W}}_r\mathbf{\tilde{T}}_{j,i}^r . 
\end{align*}
We now define gradients for updating model parameters. The gradient of objective function $J$ with respect to $\mathbf{\Lambda}_r^i$ is:
\begin{align*}
\frac{\partial J}{\partial \mathbf{\Lambda}_r^i} = \frac{\partial H(\mathcal{\tilde{W}},\mathcal{D},\mathcal{C})}{\partial \mathbf{\Lambda}_r^i} + \frac{\partial H(\mathcal{{W}},\mathcal{D},\mathcal{C})}{\partial \mathbf{\Lambda}_r^i}  
& = [\beta(-2\mathbf{\tilde{Y}}_r^\top\mathbf{\Theta}_r^i\mathbf{\tilde{Y}}_r + 2\mathbf{\tilde{Y}}_r^\top\mathbf{\tilde{Y}}_r\mathbf{\Lambda}_r^i\mathbf{\tilde{Y}}_r^\top\mathbf{\tilde{Y}}_r) \\& \hspace{-2em} + \beta (-2\mathbf{Y}_r^\top\mathbf{\Theta}_r^i\mathbf{Y}_r + 2\mathbf{Y}_r^\top\mathbf{Y}_r\mathbf{\Lambda}_r^i\mathbf{Y}_r^\top\mathbf{Y}_r)] \circ \mathbf{I}_{k_r}.
\end{align*}
The gradient of $J$ with respect to $\mathbf{W}_r$ is:
\begin{align*}
\frac{\partial J}{\partial \mathbf{W}_r} = \frac{\partial H(\mathcal{{W}},\mathcal{D},\mathcal{C})}{\partial \mathbf{W}_r} &= \sum_{i=1}^S\sum_{j=r}^{K} -4\mathbf{Y}_{r-1}^\top\mathbf{\Theta}_i\mathbf{Y}_{r-1}\mathbf{W}_r\mathbf{T}_{j,i}^r \\ & +   4\mathbf{Y}_{r-1}^\top\mathbf{Y}_{r-1}\mathbf{W}_r\mathbf{T}_{j,i}^r\mathbf{W}_r^\top\mathbf{Y}_{r-1}^\top\mathbf{Y}_{r-1}\mathbf{W}_r\mathbf{T}_{j,i}^r.  
\end{align*}
\subsection{AMSgrad update rule}
Let $g_i$ be the partial derivative of the objective function with respect to the parameter $w_i$ at $i^{th}$ iteration. Let $m_i$ and $v_i$ denote the decaying averages of past and past squared gradients, then the update rule for AMSgrad is defined as:
\begin{align*}
    m_i &= \beta_1m_{i-1} + (1-\beta_1)g_i,  \qquad
    v_i = \beta_2v_{i-1} + (1-\beta_2)g_i^2 \\
        \hat{m}_i &= \frac{m_i}{1 - \beta_1^i} \qquad
        \hat{v}_i = \text{max}(\hat{v}_{i-1},v_i) \qquad
    w_{i+1} = w_i - \frac{\eta}{\sqrt{\hat{v}_i} + \epsilon}\hat{m}_i,
\end{align*}
where $\beta_1 = 0.9$,  $\beta_2=0.99$, $\epsilon = 10^{-8}$ and $\eta = 0.1$. $\beta_1$ and $\beta_2$ are the hyperparameters in the update rules described above. These are typical values for the practical applications \cite{reddi2019convergence}. 

\begin{figure}
     \begin{subfigure}[t]{0.3\textwidth}
        \centering
        \includegraphics[width=\textwidth, valign=c]{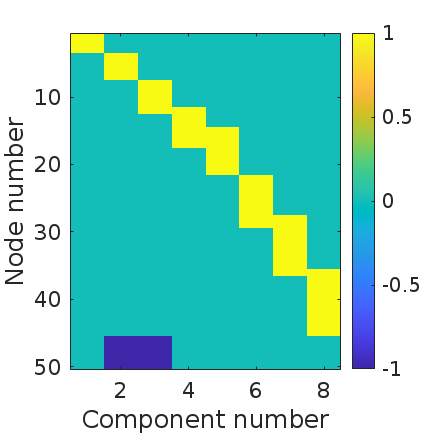}
        \caption{}
    \end{subfigure}%
    \Large{$\times$}
    \begin{subfigure}[t]{0.3\textwidth}
        \centering
        \includegraphics[width=\textwidth, valign=c]{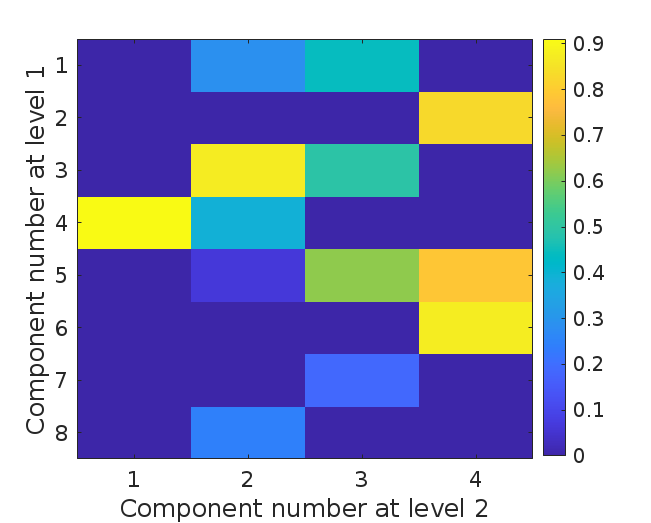} \vspace{.6em}
        \caption{}
    \end{subfigure}
    =
    \begin{subfigure}[t]{0.3\textwidth}
        \centering
        \includegraphics[width=1.1\textwidth, valign=c]{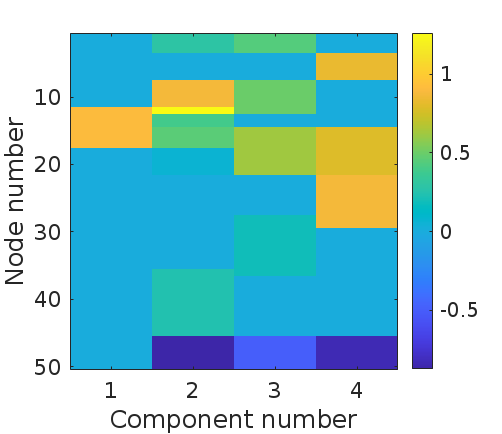} \vspace{.1em}
        \caption{}
    \end{subfigure}
    \label{fig:ground}
    \caption{(a) Visualization of ground truth components at level $1$. (b) Weight matrix used to generate components at level 2. (c) Visualization of ground truth components at level $2$. }
\end{figure}
\vspace{-3em}
\begin{table}[b!]
\caption{Accuracy on simulated dataset with $\text{Pois(0.4)}$ noise added}\label{table:simu_noise}
\centering
        \begin{tabular}{ m{2cm}P{2.3cm}P{2.3cm}P{2.3cm}P{2.3cm}  }
 \toprule
 \multicolumn{1}{c}{Method} & \multicolumn{1}{c}{$k_1=6$} & \multicolumn{1}{c}{$k_1=8$}  & \multicolumn{1}{c}{$k_1=10$} & \multicolumn{1}{c}{$k_1=12$} \\
 \midrule
hSCP & $0.798 $    & $0.779$ &   $0.739$ & $0.724$\\
Adv. hSCP & $0.804$     & $0.771$ &  $0.818$  & $0.843$ \\
ICA & $0.637 \pm 0.015 $     & $0.671 \pm 0.034$ &  $0.715 \pm 0.027$  & $0.738 \pm 0.012$ \\
NMF & $0.640 \pm 0.101 $     &$0.655 \pm 0.109$  &  $0.703 \pm 0.079$  &  $0.704 \pm 0.128$\\
Adv. NMF & $0.690\pm 0.079$     & $0.681 \pm 0.071$ &  $0.694 \pm 0.080$  & $0.682 \pm 0.088$ \\
 \bottomrule
\end{tabular}
\end{table}

\end{document}